\documentclass[]{spie}  

\usepackage{amsmath,amsfonts,amssymb}
\usepackage{graphicx}
\usepackage[colorlinks=true, allcolors=blue]{hyperref}
\usepackage{soul}
\usepackage{booktabs}
\usepackage{hyperref}
\usepackage{multirow}

\title{HALO: An Ontology for Representing and Categorizing Hallucinations in Large Language Models}

\author[a]{Navapat Nananukul}
\author[a]{Mayank Kejriwal}
\affil[a]{University of Southern California, Information Sciences Institute, Los Angeles CA, United States of America}

\authorinfo{Further author information: (Send correspondence to M.K.)\\M.K.: E-mail: kejriwal@isi.edu, Telephone: 1 310 822 1511}

\begin{document} 
\maketitle

\begin{abstract}
Recent progress in generative AI, including large language models (LLMs) like ChatGPT, has opened up significant opportunities in fields ranging from natural language processing to knowledge discovery and data mining. However, there is also a growing awareness that the models can be prone to problems such as making information up or `hallucinations', and faulty reasoning on seemingly simple problems. Because of the popularity of models like ChatGPT, both academic scholars and citizen scientists have documented hallucinations of several different types and severity. Despite this body of work, a formal model for describing and representing these hallucinations (with relevant meta-data) at a fine-grained level, is still lacking. In this paper, we address this gap by presenting the \emph{Hallucination Ontology} or HALO, a formal, extensible ontology written in OWL that currently offers support for six different types of hallucinations known to arise in LLMs, along with support for provenance and experimental metadata. We also collect and publish a dataset containing hallucinations that we inductively gathered across multiple independent Web sources, and show that HALO can be successfully used to model this dataset and answer competency questions. 
\end{abstract}


\section{Introduction}\label{intro}  

The recent advent of large language models (LLMs) \cite{lewis2019bart,openai2023gpt4,thoppilan2022lamda} and other generative AI (GenAI) systems \cite{doi:10.1126/science.abq1158,singhal2023expertlevel,zeng2022lion} has led to impressive success on a wide set of tasks, in communities ranging from natural language processing to computer vision \cite{bang2023multitask,ramesh2021zeroshot,rawte2023survey}. While the success of these models is promising, concerns have been raised about their responsible use and deployment \cite{zhang2023language}. Like other deep learning models preceding them, such models can be prone to faulty (and often, unpredictable) reasoning on relatively simple problems, and can also `make up' or \emph{hallucinate} information \cite{li2023halueval}. 

The issue of hallucinations, in particular, has raised some important concerns such as misinformation and over-reliance on the system's output, especially when the system \emph{seems} to be giving the correct answer in response to a prompt \cite{zhang2023sirens}. It has also become evident through several recently published studies \cite{rawte2023survey,ye2023cognitive} that hallucinations are not uncommon occurrences. Even `citizen scientists,' who have been playing with models like ChatGPT owing to its easy-to-use interface, have documented a wide body of hallucinations even in common Web forums.

This paper is motivated by the observation that, despite the growing number of studies (whether academic or informal), a standard vocabulary or ontology for representing hallucinations has thus far been lacking. Such an ontology, if it existed, and datasets modeled using the classes and predicates in the ontology, would serve some important hypothetical use-cases. For example, they would allow us to gain a better understanding of, and represent, different \emph{categories} of hallucinations that have been documented. 
However, without a common vocabulary, data collected on this important phenomenon would be ad-hoc and require significant manual work, making it more difficult and time-consuming to conduct systematic empirical studies. 

At the same time, any such ontology can only have practical utility if it is extensible, well-documented, and amenable to modeling both hallucination \emph{instances} and their \emph{metadata}, which include details such as what LLMs the hallucination instance was tried on, when it was tried, its provenance, and so on. Even preliminary evaluations show that all LLMs don't hallucinate equally: on average, given a set of prompts, there are clear differences between the major language models currently in circulation, as we subsequently show. 

With these motivations in mind, we propose the \emph{Hallucination Ontology} (HALO), a resource for modeling hallucination instances and their metadata. Specific contributions in this work are as follows:

\begin{itemize}
    \item We propose and describe HALO, an ontology that treats hallucinations in GenAI models as first-class citizens of study. HALO is released under an open license, follows the FAIR principles, and is intended to be sustainable and extensible in the foreseeable future, serving as an ontological resource for researchers who continue to study and gather data on hallucinations in GenAI models and LLMs.
    \item HALO was constructed following a rigorous design process and has support for multiple hallucination categories and sub-categories, metadata fields and attributes, and an attractive set of technical specifications and complete documentation. It contains two high-level modules, called the \emph{Hallucination Module} and \emph{Metadata Module}, which allows us to maintain clear separation between the (more extensible) categories representing different hallucination types, and the (more standard) concepts and properties that are better suited for capturing experimental data on hallucinations.
    \item We demonstrate the utility of HALO using a set of competency questions (CQs) that are similar to those normally studied in empirical studies of hallucination. We use a real dataset on hallucinations collected from a diverse set of sources on the Web, and show that even relatively complex CQs can be formalized as SPARQL queries using HALO concepts and properties. We verify the correctness of the answers obtained from executing these queries.  
\end{itemize}

\section{Related Work}

\textbf{Hallucinations in Generative AI (GenAI): }The advent of GenAI and large language models (LLMs) like ChatGPT and BARD mark a significant evolution from an earlier generation of models. Technically, such models are distinguished by their use of reinforcement learning in addition to deep transformer neural networks based on attention. Although deep learning models had been known to be prone to problems such as overconfidence and adversarial attacks, hallucinations have been recognized as a relatively novel class of errors and `false predictions' in the context of GenAI  \cite{yuan2022biobart,semeniuta2019accurate}. Hallucinations are not uncommon, and researchers have documented them (or errors resembling them) in a diverse range of generative tasks, such as text summarization \cite{li2023halueval}, text generation \cite{varshney2023stitch,mündler2023selfcontradictory}, multi-modal problem solving \cite{bang2023multitask}, dialogue \cite{10207581}, question-answering \cite{cui2023chatlaw,manakul2023selfcheckgpt,zhang2023language}, and language translation \cite{10.1145/3571730,guerreiro2023hallucinations}. Inevitably, as we have learned more about hallucinations as a community, their definition and scope has expanded. Much more recently, the evaluation of hallucinations in LLMs has transitioned to focus more on the distinction between factual reality, fabricated results, and the ability of LLMs to follow human instructions \cite{rawte2023survey}, as opposed to problems that can be traced back purely to faults in abstract reasoning or over-confidence. 

Documented hallucinations in LLMs tend to be domain-specific, with question-answering (QA) being a prominent example of a task that can often elicit hallucinations. Numerous datasets and benchmarks have been developed for QA over the past year \cite{zhuang2024toolqa,wei2024measuring,yang2023new}.  Additionally, efforts have been made to systematically generate hallucinations for ChatGPT in the QA domain, integrating with supervised human annotation to establish QA benchmarks \cite{li2023halueval}. The phenomenon of hallucination also extends to Large Vision-Language Models (LVLMs), where the model can hallucinate on producing an object and misidentify features that contradict textual and visual information \cite{liu2024phd}. These benchmarks are useful for task-specific prompts such as QA and text-visual generation, which motivated research in detecting and mitigating hallucinations, as illustrated in recent papers \cite{manakul2023selfcheckgpt,rawte2023survey,ji2023towards}. However, the interaction with newer or commercial LLMs like ChatGPT\cite{achiam2023gpt} often involves free-flowing text input, diverging from structured QA formats. This shift poses a challenge, as the free-form nature of interactions can lead to varied and unpredictable hallucination instances, making them harder to detect and document systematically. To address this, we are motivated in this paper to collect these instances and build a dataset and a structured framework, such as a hallucination ontology, to categorize and analyze these hallucinations systematically.

Based on performance metrics, modern LLMs are considered definitively superior compared to an earlier generation of language models like BERT. They can handle complex prompts and instructions, but this complexity has also resulted in more sophisticated hallucination patterns \cite{bubeck2023sparks}.  Researchers are already attempting to define and categorize hallucinations in LLMs in a rigorous way, but efforts are ad-hoc and do not rely on the formal machinery that is a strength of the Semantic Web \cite{huang2023survey,zhang2023sirens,jiang2024survey}. Nevertheless, such works do offer insights into hallucination causes, types, and potential solutions, even though they do not always take into account the hallucinations published in non-academic sources. 

\textbf{Ontology development and guidelines:} To the best of our knowledge, there is no existing hallucination ontology in the literature that has a formal representation. Recognizing this gap, we drew upon the Linked Open Terms (LOT) methodology \cite{POVEDAVILLALON2022104755} as our guiding framework. LOT's methodology is particularly suited to our goal of developing a robust, extensible ontology from scratch. Moreover, to connect hallucination in LLMs with existing classes and enhance interoperability, we also follow the \emph{Minimum Information to Reference an External Ontology Term (MIREOT)} guidelines \cite{mireot_article}, as subsequently discussed. Beyond hallucinations, obviously, there are many resources and decades of related work in the Semantic Web on best practices for publishing ontologies expressing novel phenomena. An exhaustive survey of this research is well beyond the scope of any reasonable work, but some recent research that is relevant to the design and development of HALO include \cite{onto1,onto2,onto3,onto4,onto5}.

\section{HALO: Design and Development}

In this section, we present the ontology design and development process for HALO. Much of this process followed the Linked Open Terms (LOT) methodology \cite{POVEDAVILLALON2022104755}, an industrial method for developing ontologies and vocabularies. The LOT methodology is structured around iterative cycles through a foundational workflow, including four phases: \textit{(i) ontological requirements specification, (ii) ontology implementation}, \textit{(iii) ontology publication}, and \textit{(iv) ontology maintenance}. The methodology also places much emphasis on the reuse of terms (ontology classes, properties, and attributes) in published vocabularies or ontologies. Before describing these four steps and how they apply to HALO, we begin by describing a use case motivating the need for an ontology like HALO.

\subsection{Use Case}

An important use case for HALO is to serve as a standard vocabulary and model for representing the many hallucinations that are being discovered and described by a broad set of individuals, not just academic scholars. Examples of hallucination can be found on the Web, especially for the more popular LLMs such as ChatGPT, published by the likes of journalists, technologist enthusiasts, and others, who  in their role as ``citizen scientists'', are experimenting organically with the LLMs. This data is valuable and may even serve an important role both in formulating interesting research questions and in helping companies and organizations improve the LLMs in further iterations, but because the data and ``experiments'' are so decentralized,  they are currently not being captured either in a central forum or using a standardized vocabulary. HALO is meant to serve the latter purpose more than the former, but we also demonstrate its feasibility for the former when we evaluate its merits in Section \ref{eval}. Some of the types of questions that HALO can help us answer include: are LLMs relatively uniform in which prompts they hallucinate on, or are there significant differences between LLMs? What kinds of hallucination categories are most common? Furthermore, given longitudinal data, can we determine if LLMs are `improving' on some categories of hallucinations, compared to others? While the last of these cannot be addressed with the data we currently have available, due to the recency of the models, we do address the first two in Section \ref{eval}.  





\subsection{Specification of Ontology Requirements}

This subsection details two types of requirements considered in HALO's construction: 

\begin{enumerate}
    \item \textbf{Functional requirements} are defined by competency questions (CQs), which contain multiple questions on different views related to GenAI hallucinations. Additionally, SPARQL queries corresponding to each CQ were created to validate HALO (Section \ref{Ontoeval}).

    \item \textbf{Non-functional requirements} for HALO include \emph{extensibility},  \emph{interoperability}, and  \emph{maintainability}.  \emph{Extensibility} is important because HALO was designed to evolve and expand as we learn more about GenAI and the different categories of hallucinations it is prone to. Hence, we followed hallucination taxonomies presented in the most recent literature on the subject, and which have a clear hierarchical structure and are easy to understand. \emph{Interoperability} is also an important feature we aimed to achieve when building HALO, by connecting each hallucination type with external classes in order to reuse existing resources from published vocabularies. Finally, we also considered \emph{maintainability} as an important design element when building HALO. In addition to following standard guidelines in the ontology's construction and design, we also created a pipeline to extend the hallucination categories and broaden the scope of external classes. This pipeline includes Google Forms, Sheets, and a Python-based program to convert the data into RDF/XML format.

\end{enumerate}

\subsection{Ontology Implementation}

In developing HALO, we aimed to closely adhere to the general use cases and motivations we had previously introduced in Section \ref{intro}. For example, we had stated therein that supporting a broad set of experimental studies involving actual hallucinations was an important reason for proposing a standard vocabulary. To conceptualize this need, HALO was designed to contain two modules: (1) the \emph{Metadata Module} and (2) the \emph{Hallucination Module}. We created the initial conceptual model using Chowlk\footnote{\url{https://chowlk.linkeddata.es/index.html}} and implemented the OWL version using Protégé \cite{10.1145/2757001.2757003}. From a socio-technical perspective, the design and developmental process of HALO involved iterative internal discussions aimed at refining the ontology to implement the requirements (both functional and non-functional) noted earlier. These discussions spanned both data collection and literature review strategies for ensuring that, even in its first version, HALO contained a minimal set of hallucination categories and external entities for representing the hallucinations that people had already discovered. Through the review of the literature, we also sought to align our categories with hallucination terminologies to ensure that we were using commonly used terms. This ensured that our ontology remained up-to-date with the latest findings and best practices in the growing literature in this area. Finally, we evaluated HALO followed the standard LOT process by asking for feedback and reviewing external knowledge, not just in the academic literature but also informal sources on the Web. Following these guidelines, we created competency questions and corresponding SPARQL queries to fully validate the ontology. Using this iterative process, we were able to refine HALO to its current version, the OWL version of which is provided in the resource links included with this paper.

\subsection{Linking External Vocabularies}

HALO is created based on two existing primary vocabulary sources: (1) hallucination categories and (2) external classes/entities. As a key objective is to ensure interoperability, we took the following systematic approach:

\textbf{Identify Top-Level Hallucination Categories}: As a first step, we identified primary categories of hallucinations observed in LLMs through a careful review of the literature \cite{huang2023survey,zhang2023sirens}. These categories form the foundational classes in HALO's \emph{Hallucination Module}.

\textbf{Link Hallucinations with External Classes}: We planned to import related vocabularies to make HALO as comprehensive as possible, but because not all classes and properties in such vocabularies (like FOAF) are relevant, and to avoid considerable overhead of full imports, we
adhered to the \emph{Minimum Information to Reference an External Ontology Term (MIREOT)} guidelines, which suggest that the essential information needed to reference an external concept includes the ontology namespace, the URI of the specific term, and the URI for the term's superclass \cite{mireot_article}. By following MIREOT, we selectively integrated relevant external concepts without also including superfluous terms. We also established connections between external classes that might appear in actual hallucination instances e.g., when an LLM hallucinates a fact about a person, we connect \textit{foaf:person} to our hallucination module, ensuring that the semantics of the hallucination are reliably represented in reused vocabulary terms.

\subsection{Ontology Publication and Maintenance} \label{onpub}

HALO is published under the open CC-BY 4.0 license in two formats: (1) a machine-readable (OWL) file following ontology-publication best practices \cite{w3best}, with the files  available both in a GitHub repository and a Zenodo repository (Section \ref{sec:tech} provides the links and details); (2) a human-readable document that allows for more conceptual debate on how to represent hallucination categories and metadata. We documented HALO using Widoco\cite{10.1007/978-3-319-68204-4_9}, which generates a series of interlinked HTML pages offering a human-readable version of HALO, derived directly from its content. This documentation is made publicly accessible online through a persistent identifier, allowing us to fulfill  FAIR's \emph{Findability} (specifically, F1, F2, and F4) requirements. 

Concerning ontology maintenance, we plan to keep an active GitHub issue tracker, with the repository open for users to submit issues and suggestions. In particular, we solicit discovery of bugs through this avenue, and pitfalls that we may have missed in this version, although we also intend the tracker to enable external collaboration, requests for extension and general feedback.

To further maintain and promote the ontology, we will also be launching a Google form that will allow users or researchers to report actual hallucination instances that we will model using the ontology. This form, although not directly related to the development or extension of the ontology itself, is expected to facilitate an important use case that motivated the development of HALO (collecting and analyzing experimental data on hallucinations in a standard and relatively automated way). Particularly, we  hope to periodically review submitted hallucinations to determine whether the ontology should be extended with more hallucination categories. 

\section{HALO: Hallucination Ontology}

\begin{figure*}[ht]
    \centering\footnotesize
    \includegraphics[width=0.75\textwidth]{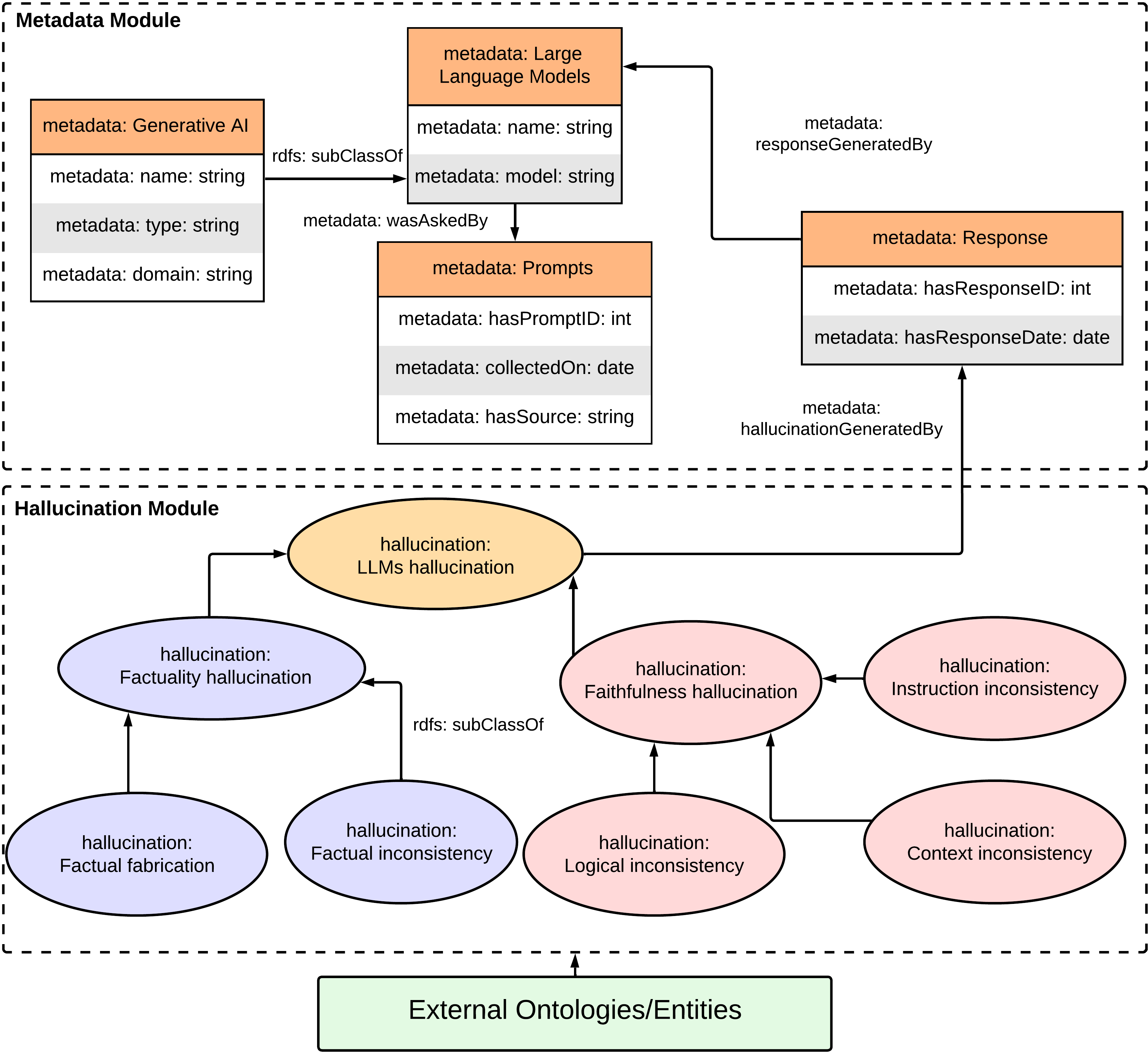}
    \caption{A visualization of HALO, including the \emph{Metadata} and \emph{Hallucination} modules that describe it at the highest level and are further articulated in the main text.} 
    \label{fig:metadata}
\end{figure*}

This section describes the architecture and elements of the HALO, which is structured into two principal modules: (1) the \emph{Hallucination} module, and (2) the \emph{Metadata} module. We begin by defining (Section \ref{hallCat}) the different hallucination categories that HALO supports and that evidence has shown that the popular LLMs (such as ChatGPT and BARD) exhibit. The categories and definitions below provide the foundation for HALO and the phenomena it aims to capture. Figure \ref{fig:metadata} schematizes the ontology in a visually accessible manner. As discussed earlier, the ontology has been formally published and released under an open-source license as an OWL file, and is designed to be extensible. Following a discussion of the hallucination categories and sub-categories, we detail the two different modules in Sections \ref{hallHall} and \ref{hallMeta}.

\subsection{Hallucination Categories}\label{hallCat}

Even in the short time that LLMs have been exposed to the public, several studies have found that hallucinations that they are prone to are not all equal \cite{huang2023survey,zhang2023sirens}. Guided by this literature, we chose to include two main hallucination categories (with subcategories) into HALO. Eventually, we believe that more hallucination categories and subcategories might be discovered, which is the reason that the hallucinations below are sub-classes of a broader \emph{hallucination} class.

\textbf{1. Factuality Hallucinations} refers to errors in LLMs outputs by misrepresenting real-world facts. These errors manifest in two distinct ways:

\begin{itemize}

  \item \textbf{Factual Inconsistency} occurs when LLMs produce answers or responses that reference real-world facts but include contradictions or misinformation. 
  
  \textit{Example: LLMs incorrectly identify ``Yuri Gagarin'' as the first person to land on the Moon, a clear contradiction of established historical facts.}
  
  \item \textbf{Factual Fabrication} refers to LLMs producing responses with supporting evidence that cannot be verified by established real-world information.
  
  \textit{Example: An LLM fabricates a detailed historical narrative about dragons existing on Earth despite no empirical evidence or historical records supporting such a claim.}
\end{itemize}

\textbf{2. Faithfulness Hallucination} refers to LLM outputs that fail to align faithfully with user instructions, when provided with context, including well-known logical rules based in objective fields like math or physics, for example. We further sub-categorize it into three subtypes:

\begin{itemize}
  \item \textbf{Instruction Inconsistency} arises when LLM outputs do not follow the user's prompt instructions.
  
  \textit{Example: Users intend to request a translation of a specific sentence, but the LLM erroneously performs a question-answering to that sentence instead of translating it.}

  \item \textbf{Context Inconsistency} occurs when the LLM’s output is not aligned to the user’s provided contextual information. 

  \textit{Example: LLMs provided an answer about great food for dinner when a user asked about breakfast, and it clearly stated in the users' prompt}

  \item \textbf{Logical Inconsistency} occurs when an LLM's output exhibits internal logical contradictions, often observed in reasoning tasks\footnote{There is some controversy whether artifacts like logical inconsistencies are `hallucinations' or just ordinary drops in performance. No AI-based reasoning system is known to have perfect performance, given non-trivial natural language inputs. However, there is reason to believe that the errors produced by LLMs is grounded, not in faulty reasoning, but faulty interpretation of inputs or context, which better resembles hallucinatory phenomena. In this paper, we err on the side of caution and take a reasonably broad view of hallucinations.}. 

  \textit{Example: LLM might correctly perform mathematical steps, but then conclude with an incorrect final answer}
  
\end{itemize}

\subsection{Hallucination Module}\label{hallHall}

The \emph{Hallucination} module contains classes for each of the categories and sub-categories defined in Section \ref{hallCat} and is the main module in HALO. A key feature of the module is its integration with terms from existing vocabularies, in adherence to the Linked Data standards. Below, we discuss the two important features of the module.

\textbf{Hierarchical structure:} The hallucination categories and sub-categories we defined earlier are best encoded as a taxonomy, using appropriate \textit{owl: subclassOf} relations. This is also shown in Figure \ref{fig:metadata}. In line with the categorization provided earlier, there are two main branches from the major \textit{LLMsHallucination} root class: (1) \textit{factualityHallucination} which comprises two sub-classes \textit{FactualFabrication} and \textit{factualInconsistency} (2) \textit{faithfulnessHallucination} which comprises of three sub-classes \textit{logicalInconsistency}, \textit{instrctionInconsistency}, and \textit{contextInconsistency}. Note also that the taxonomy, laid out using a relatively simple structure, ensures that the ontology can be expanded as new hallucination categories and sub-categories emerge from the intense research that is being conducted at the present moment on LLMs. 

\textbf{Instances} within this module correspond to specific hallucinated prompts and responses identified by a researcher, or through mechanisms like citizen science. As we show in Section \ref{eval}, when we evaluate the merits of HALO using a real-world hallucination dataset collected from multiple sources from the Web, these instances often involve real-world entities such as individuals' names, geographic locations, and documented events, varying according to the user-generated prompts. Examples of external classes we connected to this module include {\textit{foaf:person}, \textit{foaf:document}, \textit{schema:date}, \textit{schema:schorlarlyArticle}, etc}. 

Finally, the ontology also provides support for recording the metadata and provenance of the different hallucinations that can be represented as instances of the hallucination classes and sub-classes (Section \ref{hallMeta}). 



\subsection{Metadata Module}\label{hallMeta}

The \emph{Metadata} module represents the core concept and properties of the prompts and answers we gathered. We visually represent its structure in the top pane in Figure \ref{fig:metadata}. The module comprises four primary concepts: \textit{GenerativeAI}, \textit{LargeLanguageModel}, \textit{LLMsPrompt}, and \textit{LLMsAnswers}, each briefly described below: \begin{enumerate}

\item \textit{GenerativeAI} describes broadly the types of AI agents that showed hallucinations. The instance of this class can be an AI agent from any field such as text generation models, image generation models, or LLMs. Since our initial experiments are tested on LLMs, the \textit{GenerativeAI} class has only one subclass (\textit{LargeLanguageModel}) for the current version of HALO. 

\item \textit{LargeLanguageModel} describes instances of LLMs that exhibit hallucination, including names and model versions.

\item \textit{LLMsPrompt} instantiates the metadata of the collected prompts, with attributes such as (1) \textit{hasPromptID}, an internal identifier for tracking prompts; (2) \textit{CollectedOn}, the date of prompt acquisition; and (3) \textit{hasSource}, the internet source we found the prompt. 

\item \textit{LLMsAnswer} encapsulates metadata about the LLMs' responses, featuring (1) \textit{hasAnswerID} for internal answer tracking and (2) \textit{hasAnswerDate} for recording the date of the response. Additionally, \textit{LLMsAnswer} serves as a bridge between the Metadata and Hallucination modules, linked by the relation \textit{metadata: hallucinationGeneratedBy}, indicating that an LLM's answer we add into our ontology generated a specific hallucination type. 
\end{enumerate}


\section{Technical Specifications and Availability}\label{sec:tech}

We summarize HALO's technical specifications augmenting those mentioned in Section 3, including \emph{interoperability, indexing and availability, logical correctness and validation}, and \emph{documentation}.

\textbf{Interoperability:} We imported seven external classes into HALO, as one of our goals in designing HALO is to represent hallucinations that involve real-world entities and concepts, which are already present in vocabularies like FOAF. Note that the core implementation of HALO is OWL-based in order to achieve FAIR's Interoperability (I1) principle.






\textbf{Indexing and availability:} We release the ontology publicly\footnote{\url{https://bitly.ws/34Sdp}}, under a permissive CC BY 4.0 International License\footnote{\url{https://creativecommons.org/licenses/by/4.0/}}, as well as in Zenodo\footnote{\url{https://zenodo.org/records/10279463}}. HALO can be browsed using any web browser, and using visualization tools such as WebVOWL. It can also be accessed on GitHub\footnote{\url{https://github.com/navapatn/halo-ontology}}.

\textbf{Logical correctness and validation:} We validated HALO against a known ontology scanner tool called \emph{OOPS! Ontology Pitfall Scanner} \cite{poveda2014oops}. This tool verifies 29 common pitfalls in ontology development, including creation of polysemous elements, synonyms as classes, and so on. The tool notifies users of three critical levels based on the number of pitfalls and the impacts on the ontology. HALO passed the evaluation with zero pitfalls using this evaluation tool.

\textbf{Documentation:} We documented HALO using the \emph{Widoco wizard} \cite{10.1007/978-3-319-68204-4_9} as mentioned in subsection \ref{onpub}. The documentation\footnote{\url{https://bitly.ws/34L6U}} aims to provide a comprehensive explanation to broad classes of users, not just domain-specific researchers. The documentation is fully accessible online, and a live visualization of the ontology is also included therein using the WebVOWL plugin, as mentioned earlier.


\section{Evaluation}\label{eval}

\subsection{Hallucination Dataset and Sources} \label{dataset}

In our effort to model and evaluate hallucination instances, we contribute by constructing a compact dataset specifically designed to document hallucinatory responses. This dataset serves as an initial repository and facilitates the evaluation of HALO's competency. We compiled 40 prompts known to induce hallucinatory responses in LLMs from a diverse set of Web sources. The prompts span multiple document types, including Reddit posts, news articles from The New York Times, and reports from Opengov, among others. During this collection process, we also compiled information on applicable metadata, such as the specific LLM known to have produced the hallucination, date when we found the hallucination, Web source link, name, and type (e.g., blog post, newspaper article). We tabulate descriptive statistics on this collected data in Table \ref{tab:additional_metadata_statistics}.

\begin{figure*}[ht]
    \centering\footnotesize
    \includegraphics[width=\textwidth]{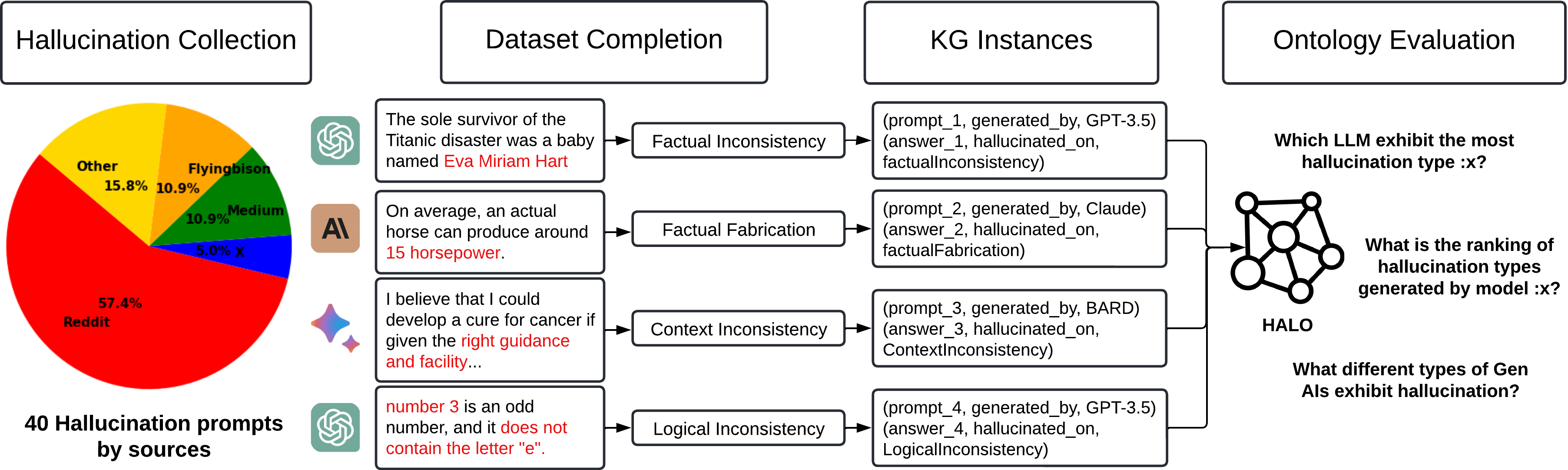}
    \caption{An illustration of the complete workflow for evaluating HALO. Hallucination instances are collected from multiple sources on the internet, typically as response examples from a single LLM. As described in Section \ref{dataset}, each prompt is executed across three different LLMs to complete the dataset, and instances are labeled as hallucinations when identified. Only the instances that exhibit hallucinations are then used to create Knowledge Graph (KG) instances, which are subsequently integrated into HALO for evaluation through SPARQL queries.} 
    \label{fig:data-process}
\end{figure*}

\begin{table}[ht]
\caption{Descriptive statistics and representative examples from the hallucination dataset that was collected for evaluating HALO}
\centering
\begin{tabular}{|p{4.8cm}|p{1.5cm}|p{5.3cm}|}
\hline
\textbf{Statistics} & \textbf{Count} & \textbf{Example} \\ \hline
Number of Prompts & 40 & [How many games did the New Jersey Devils win in 2014] \\ \hline
Number of Sources & 9 & [Reddit] \\ \hline
Number of Metadata Collected & 5 & [Source, Source\_link, Date, etc.]  \\ \hline
Number of Document Types & 4 & [Post] \\ \hline
Date found & - & [April 1, 2023]   \\ \hline
\end{tabular}
\label{tab:additional_metadata_statistics}
\end{table}

When studying hallucinations in LLMs, an important analytical objective is to assess whether hallucinations \emph{persist} across LLMs i.e., if a prompt led to a hallucination in ChatGPT, does it also lead to a hallucination in BARD, more often than not? HALO can enable us to pursue such objectives if the dataset contains a diverse set of prompts, hallucinations, and LLMs. All hallucinations in the dataset that we described earlier were originally tested on one of three different LLMs: ChatGPT (GPT-3.5 version), BARD, and Claude. At present, these models represent the forefront in language-based GenAI research and are briefly described below:


\textbf{ChatGPT (v3.5)}\cite{chatgpt} is a variant of the GPT (Generative Pretrained Transformer) family of generative language models developed by OpenAI, and has become widely known for its conversational abilities and its general knowledge.

\textbf{BARD}\cite{bard} is an LLM developed by Google AI. This model utilizes a transformer-based encoder-decoder architecture, incorporating a novel attention mechanism that allows it to effectively capture long-range dependencies within the input text.

\textbf{Claude}\cite{claude} is an AI assistant developed by Anthropic. Similar to other LLMs, it was also designed to learn from human feedback and conversations, and to be more introspective about admitting its mistakes, refusing inappropriate requests, and discussing options that are more aligned with human ethics.

\begin{figure*}[ht]
    \centering\footnotesize
    \includegraphics[width=\textwidth]{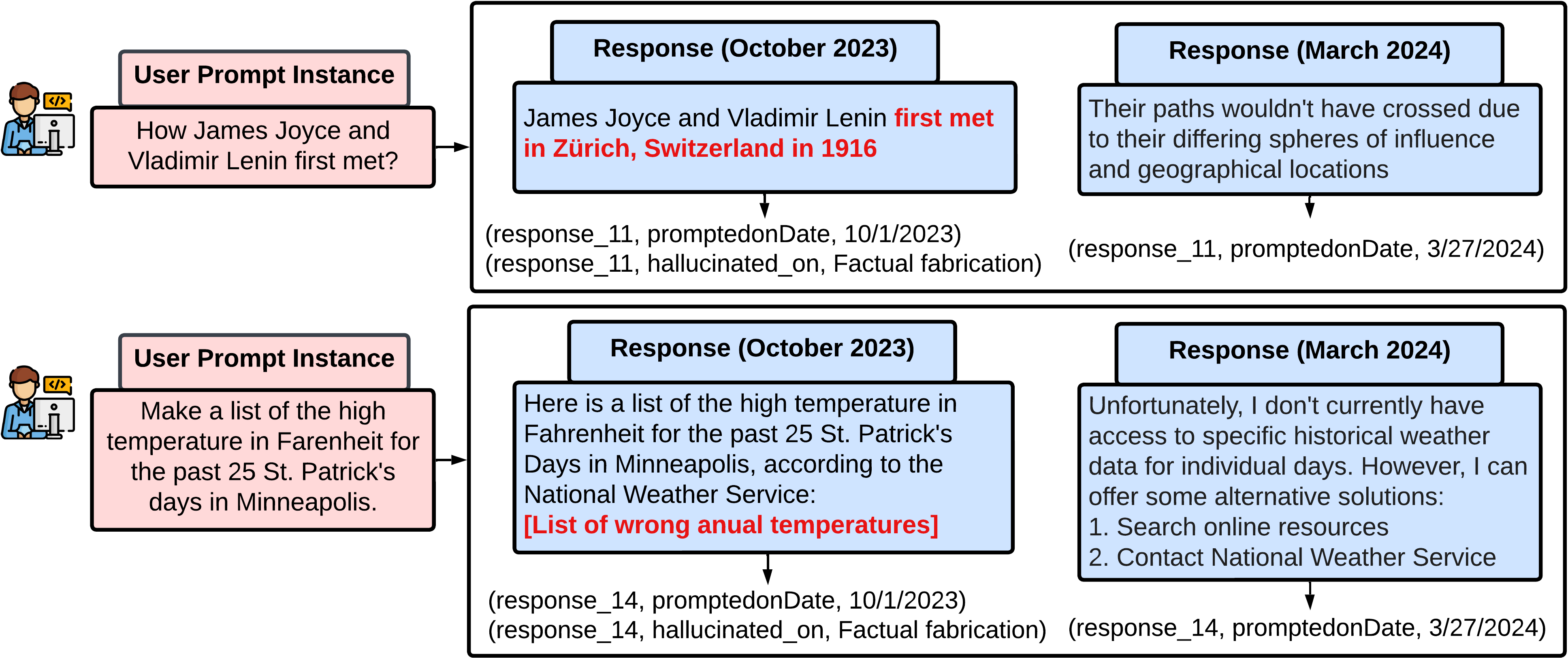}
    \caption{An illustrative example showing BARD's responses that hallucinated when prompted in October 2023, but refused to answer in March 2024.} 
    \label{fig:gemini}
\end{figure*}

Because the original dataset we collected did not evaluate all prompts on all three LLMs, to showcase HALO's utility, we first `completed' the dataset by running all 40 prompts to all three LLMs. Figure \ref{fig:data-process} illustrates the complete process from dataset collection to HALO evaluation. We ran 40 prompts in both October 2023 and March 2024, not only to complete the dataset but also demonstrate that HALO can be used to document changes to hallucination instances over time. This resulted in 120 data points for each run, which yields a total of 240 data points. As shown in Table \ref{tab:combined_statistics}, the data from October 2023 suggests variations in both the frequency and nature of hallucinations across different LLMs, despite the dataset's relatively small size.
A cursory analysis shows that, of the 40 prompts tested, BARD had the highest hallucination rate at 62.5\%, followed by ChatGPT (42.5\%) and Claude (40.0\%). Particularly, BARD's inability to refuse to respond to user prompts might have contributed to its higher rate of hallucinations. In March 2024, the results for GPT-3.5 and Claude remained consistent with results from October 2023, while BARD exhibited a slight improvement, notably in refusing to answer questions on which it has no information. Figure \ref{fig:gemini} shows prompt examples where BARD used to hallucinate in October 2023, but improved its answer in later responses.

\begin{figure*}[ht]
    \centering\footnotesize
    \includegraphics[width=0.8\textwidth]{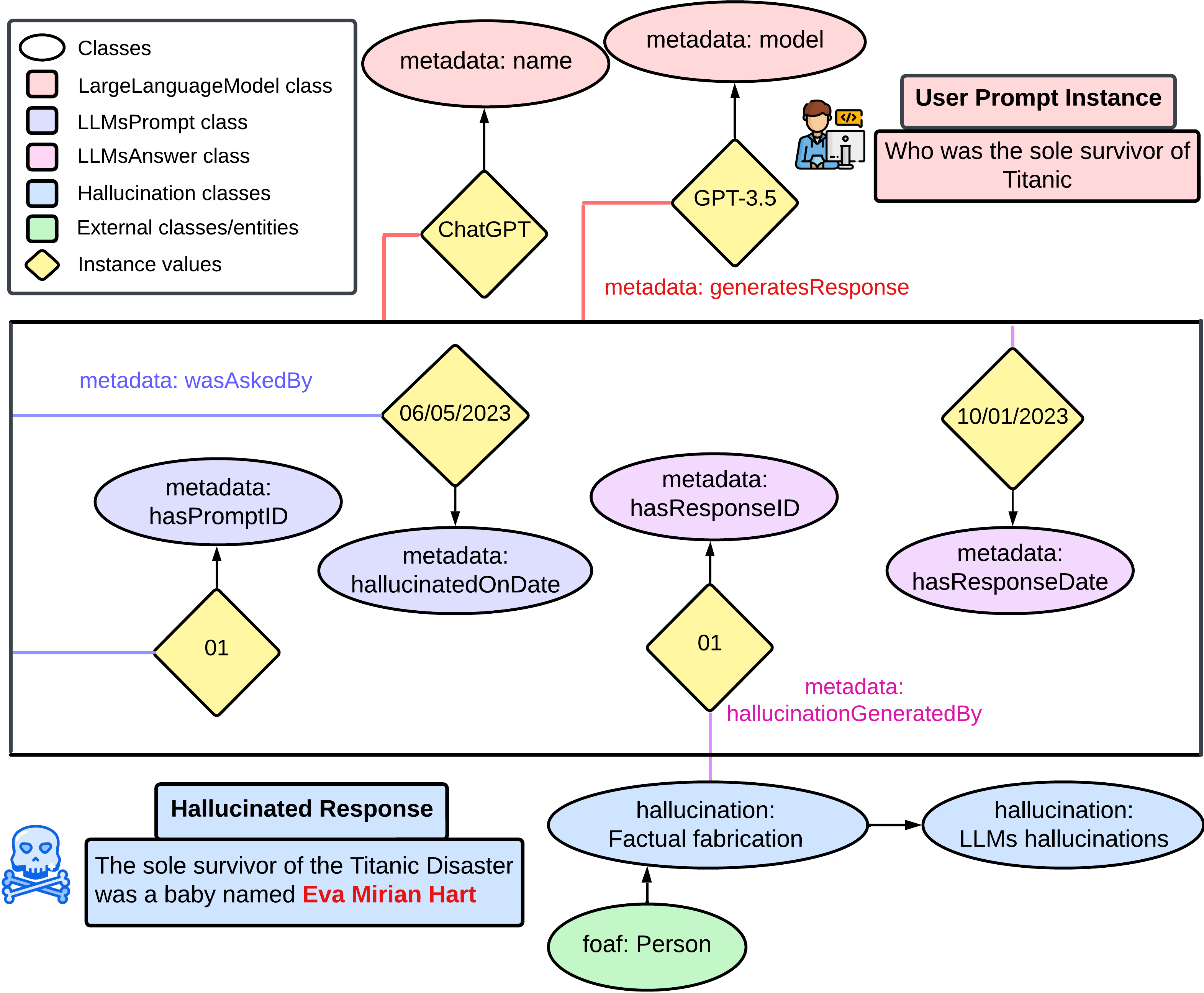}
    \caption{An illustrative example of a hallucination modeled by HALO (in KG manner) using one data point from the dataset.} 
    \label{fig:kg-1}
\end{figure*}

After completing the dataset, we generated knowledge graph (KG) instances for entries exhibiting hallucinations. Each KG instance includes all previously mentioned attributes and metadata, enhanced with hallucination labels corresponding to the category of hallucination identified in the responses. Figure \ref{fig:kg-1} illustrates a KG representation of a data point from HALO. This served two purposes: first, it allowed us to verify that a hallucination was indeed a hallucination on the model on which it was documented; second, it allowed us to pursue the stated objective through a competency query (CQ), as subsequently described.

\begin{table}[ht]
\caption{Validation statistics on hallucination experiments (using the dataset described in Table \ref{tab:additional_metadata_statistics}) on the three LLMs.}
\centering
\begin{tabular}{|p{3.0cm}|p{1.6cm}|p{1.6cm}|p{1.6cm}|p{1.6cm}|p{1.6cm}|p{1.6cm}|}
\hline
& \multicolumn{3}{c|}{\textbf{March 2024}} & \multicolumn{3}{c|}{\textbf{October 2023}} \\ \cline{2-7} 
\textbf{Statistic} & \textbf{GPT-3.5} & \textbf{BARD} & \textbf{Claude} & \textbf{GPT-3.5} & \textbf{BARD} & \textbf{Claude} \\ \hline
\# Hallucinations & 17 & 25 & 16 & 16 & 22 & 16 \\ \hline
\# Correct answers & 13 & 15 & 13 & 14 & 15 & 13 \\ \hline
\# Refuse to answer & 10 & 0 & 11 & 10 & 3 & 11 \\ \hline
Hallucination Rate & 42.5\% & 62.5\% & 40.0\% & 40.0\% & 55.0\% & 40.0\% \\ \hline
\end{tabular}
\label{tab:combined_statistics}
\end{table}

\subsection{Ontology Evaluation} \label{Ontoeval}

We evaluated HALO following the guidelines proposed in \cite{POVEDAVILLALON2022104755}. Our primary objective in this section is to show that HALO is comprehensive enough to model a range of observed hallucination types and can support a relatively rich set of analytical queries. For the latter, we devised a set of competency questions (CQs), shown in Table \ref{table:competency-questions}.

\begin{table}[ht]
\centering
\caption{A sample of the competency questions (CQs).}
\setlength\tabcolsep{12pt} 
\begin{tabular}{|p{0.8cm}|p{9.8cm}|}
\toprule
Id & Question Text \\
\midrule
CQ1 & What different types of generative AIs exhibit hallucination? \\
CQ2 & What are the different types of hallucination that occurred by large language models :x? \\
CQ3 & What is the ranking of hallucination types generated by model :x? \\
CQ4 & Which large language model exhibits the most hallucination type :x? \\
CQ5 & What pair of LLMs shared the same type of hallucinations the most? \\
\bottomrule
\end{tabular}
\label{table:competency-questions}
\end{table}

\begin{table}[ht]
\centering
\caption{Sample test cases for evaluating HALO on CQs 2, 3, and 4.}
\setlength\tabcolsep{12pt} 
\begin{tabular}{cp{3cm}p{3cm}p{3cm}}
\toprule
CQ & Input(s) & Expected Result(s) \\
\midrule
CQ2 & GPT-3.5 & FF, FI	LI, CI \\
CQ3 & BARD & (1) FI, (2) FF, (3) CI, (4) II, (5) LI   \\
CQ4 & FactualityFabrication & BARD \\

\bottomrule
\end{tabular}
\label{table:sample-test-cases}
\end{table}

Given these CQs, the evaluation is predicated on showing that HALO can be used to provide accurate answers to them. For example, based on the dataset, we would expect the response to CQ1 to be  that \emph{LLMs} (which are sub-types of GenAI) exhibit hallucination, and the response to CQ2 to be the \emph{Factuality} hallucination type  is the most frequent type to be present. To conduct the actual evaluation, as is standard in other related work \cite{10.1007/978-3-031-33455-9_26,10.1007/978-3-031-33455-9_23}, we first translate the CQs (expressed informally in natural language as shown in Table \ref{table:competency-questions}) into formal SPARQL queries framed using the concepts and properties in HALO. We note that, because HALO contains both a \emph{Hallucination} and \emph{Metadata} module, it contains the necessary ontological elements for modeling the CQs as equivalent SPARQL queries. 

Table \ref{table:sample-test-cases} presents a selection of the specific test cases used to verify HALO based on the CQs for the verification of HALO. We found that HALO achieved perfect scores across these deployed test cases, showing (to a reasonable first approximation) not only that the ontology is rich enough for us to formally express the types of CQs in Table \ref{table:competency-questions} as SPARQL queries, but also that the current version of HALO has been correctly implemented and is giving expected results for those queries. We also conducted further testing to verify the latter of these claims, and to ensure to the best of our ability that the public version of the ontology does not have implementation or logical errors.

\begin{figure*}[ht]
    \centering\footnotesize
    \includegraphics[width=0.9\textwidth]{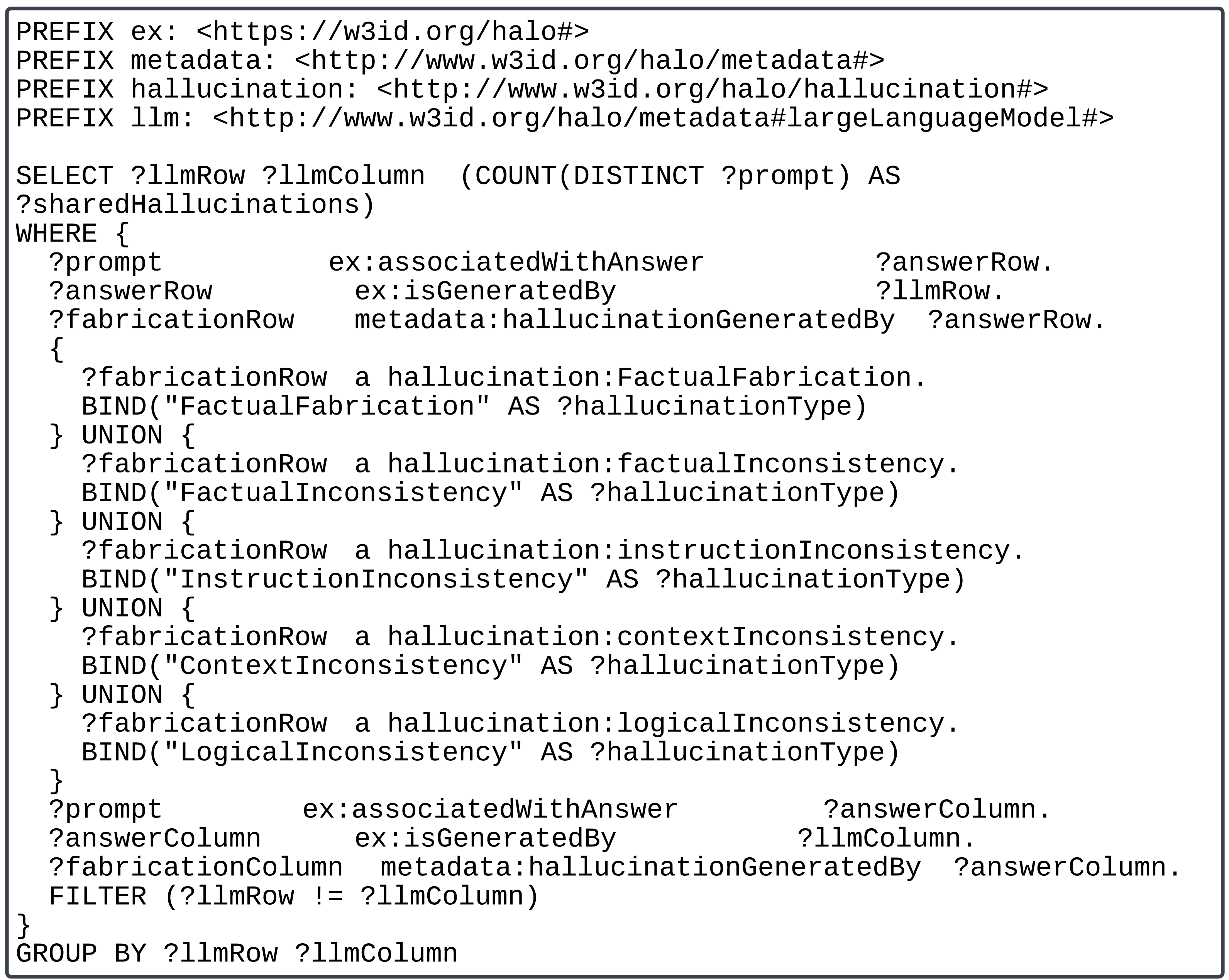}
    \caption{An example of SPARQL used to validate CQ5} 
    \label{fig:spaql}
\end{figure*}

\begin{table}[ht]
\centering
\caption{The results from executing the SPARQL query corresponding to CQ5, expressed as a confusion matrix. Since the query only counts hallucinations common to distinct (pairs of) LLMs, the diagonal entries do not apply.}
\setlength{\tabcolsep}{8pt} 
\begin{tabular}{lccc}
\toprule
          & \textbf{GPT-3.5} & \textbf{Claude} & \textbf{BARD} \\
\midrule
\textbf{GPT-3.5}   & -       & 9      & 12   \\
\textbf{Claude}    & 9       & -      & 10   \\
\textbf{BARD}      & 12      & 10     & -    \\
\bottomrule
\end{tabular}

\label{table:confusion-matrix}
\end{table}

 Due to space limitations, we only show the most complex SPARQL query (shown in Figure \ref{fig:spaql}) used to verify CQ5, with execution results tabulated in Table \ref{table:confusion-matrix}. We ran this query over 60 hallucination instances and consequently populated the confusion matrix, which shows the pairwise association between different models' (GPT-3.5, Claude, and BARD) hallucinations. The first row shows, for example, that there are nine common prompts on which GPT-3.5 and Claude both hallucinate. Overall, the successful execution of this query on HALO enables this type of analysis that enables us to compare different LLMs on their ability to hallucinate. In the future, as more instances of hallucination classes in the \emph{Hallucination Module} are collected, and the dataset used in this section is augmented, it becomes possible to conduct analyses of even broader scope using a set of CQs modeled in the same vein as CQ5.

\section{Conclusion and Future Work}

Since the release of ChatGPT, DALL-E and other such models, there has been enormous progress on GenAI problems across domains and applications. However, the complexity of these deep learning systems has also given rise to unique problems, such as hallucinations. Over the last few months, both academics and ordinary people have discovered and documented interesting sets of hallucinations when `playing' with these models. A systematic study of hallucinations remains uncommon, and to our knowledge, a standard vocabulary for expressing and analyzing them is still lacking. In this paper, we proposed HALO, a hallucination ontology that provides rich facilities for modeling six diverse types of hallucinations (themselves categorized into two larger categories), and that also provides support for modeling experimental metadata on actual hallucinations. We demonstrated the utility of HALO using a set of practical competency queries that would likely be of interest to researchers and practitioners alike.    

There are many avenues for future research. In releasing HALO publicly, and adhering to time-established principles (such as using external vocabularies where possible), we hope to sustain and extend the ontology in the long-term future. Without a doubt, more categories and sub-categories of hallucinations will have to be added to HALO both as the existing GenAI systems and LLMs become more powerful (and consequently, hallucinate in ways we have not observed yet), and as we study existing LLMs in more detail. We also hope to add support for longitudinal experimental studies to HALO, and to supplement the competency queries with documented hallucinations from other (e.g., non-language-based) GenAI models.

\bibliography{report} 
\bibliographystyle{spiebib} 

\end{document}